\documentclass[conference]{IEEEtran}

\IEEEoverridecommandlockouts  %

\usepackage{times}

\usepackage[numbers,sort&compress]{natbib}
\usepackage{multicol}
\usepackage[bookmarks=true]{hyperref}
\usepackage{graphicx}
\usepackage{amsmath}
\usepackage{booktabs}
\usepackage{amssymb}%
\usepackage{pifont}%
\usepackage{algorithm}
\usepackage{algpseudocode}
\newcommand{\cmark}{\ding{51}}%
\newcommand{\xmark}{\ding{55}}%

\begin{document}

\title{\LARGE \bf \texttt{frax}: Fast Robot Kinematics and Dynamics in JAX}

\author{Daniel Morton and Marco Pavone%
\thanks{Daniel Morton was supported by a NASA Space Technology Graduate Research Opportunity}%
\thanks{Daniel Morton and Marco Pavone are with the Departments of Mechanical Engineering and Aeronautics \& Astronautics, Stanford University, Stanford, CA 94305.
        {\tt\small \{dmorton, pavone\}@stanford.edu}}%
}

\maketitle

\begin{abstract}
In robot control, planning, and learning, there is a need for rigid-body dynamics libraries that are highly performant, easy to use, and compatible with CPUs and accelerators. While existing libraries often excel at either low-latency CPU execution or high-throughput GPU workloads, few provide a unified framework that targets multiple architectures without compromising performance or ease-of-use. To address this, we introduce \texttt{frax}, a JAX-based library for robot kinematics and dynamics, providing a high-performance, pure-Python interface across CPU, GPU, and TPU. Via a fully-vectorized approach to robot dynamics, \texttt{frax} enables efficient real-time control and parallelization, while supporting automatic differentiation for optimization-based methods. On CPU, \texttt{frax} achieves low-microsecond computation times suitable for kilohertz control rates, outperforming common libraries in Python and approaching optimized C++ implementations. On GPU, the same code scales to thousands of instances, reaching upwards of 100 million dynamics evaluations per second. We validate performance on a Franka Panda manipulator and a Unitree G1 humanoid, and release \texttt{frax} as an open-source library.

\end{abstract}

\IEEEpeerreviewmaketitle

\section{Introduction}
\label{sec:introduction}

As robots move faster and out of purely kinematic, quasi-static settings, an understanding of dynamics is critical for computing or learning real-world-feasible planners, controllers, and safety filters. Many strong libraries exist for computing robot dynamics, with Pinocchio \cite{carpentier2019pinocchio} typically noted as the fastest, though closely followed by alternatives \cite{Felis2016, frigerio2016robcogen, rigidbodydynamicsjl, neuman2019benchmarking}. Similarly, physics engines and simulators are often used in control and planning to extract dynamics properties, on CPU \cite{todorov2012mujoco, drake, coumans2021} and GPU \cite{mittal2025isaaclab, newton2025physics, Genesis, brax2021github}. 

There have also been a number of fast libraries for motion planning and inverse kinematics, including those that use GPU acceleration \cite{curobo_report23}, single instruction / multiple data (SIMD) operations \cite{vamp_2024}, or native JAX implementations \cite{kim2025pyroki}. Others focus on differential inverse kinematics, with Python interfaces to compiled libraries \cite{pink, Zakka_Mink_Python_inverse_2026}.

Recent work has also focused extensively on analytical and accelerated computation of gradients of rigid body dynamics algorithms \cite{carpentier2018analytical, plancher2021accelerating, plancher2022grid}, with applications to nonlinear trajectory optimization, MPC, and optimal control more broadly \cite{mastalli20crocoddyl, aligator, tsid}. 

Most of the leading dynamics libraries are written in C++, and while they often provide Python bindings, this can lead to added development challenges and overhead when integrating into Python-based workflows. The best libraries on CPU cannot be directly run on GPU, and vice-versa. Likewise, even if large-scale parallelization on GPU is not needed, C++-based libraries often lead to difficulties when using automatic differentiation (AD) in Python, whether in machine learning, robot planning, or control \cite{morton2025oscbf}. An ideal library would run across devices, be easy to prototype with, deploy efficiently in high-performance settings, and be compatible with AD libraries in Python, such as JAX.

\subsection{Contributions}

To address this, we present \texttt{frax}, a pure-Python library for fast computation of robot kinematics and dynamics. Due to its vectorized JAX implementation of rigid-body dynamics algorithms, \texttt{frax} natively supports just-in-time (JIT) compilation, compatibility with CPU, GPU, and TPU, and automatic differentiation. In comparisons to common dynamics libraries with Python interfaces (Pinocchio, MuJoCo), \texttt{frax} is the fastest library for writing inverse-kinematic and inverse-dynamic controllers, coming close to matching the raw performance of optimized C++. The \textit{same} code can then be deployed across thousands of parallel instances, meeting or exceeding the performance of libraries for GPU-based simulation (MJX, BRAX). \texttt{frax}'s codebase can be found at {\small\texttt{\url{https://github.com/danielpmorton/frax}}}. %

\begin{figure}[t]
    \centering
    \includegraphics[width=0.9\linewidth]{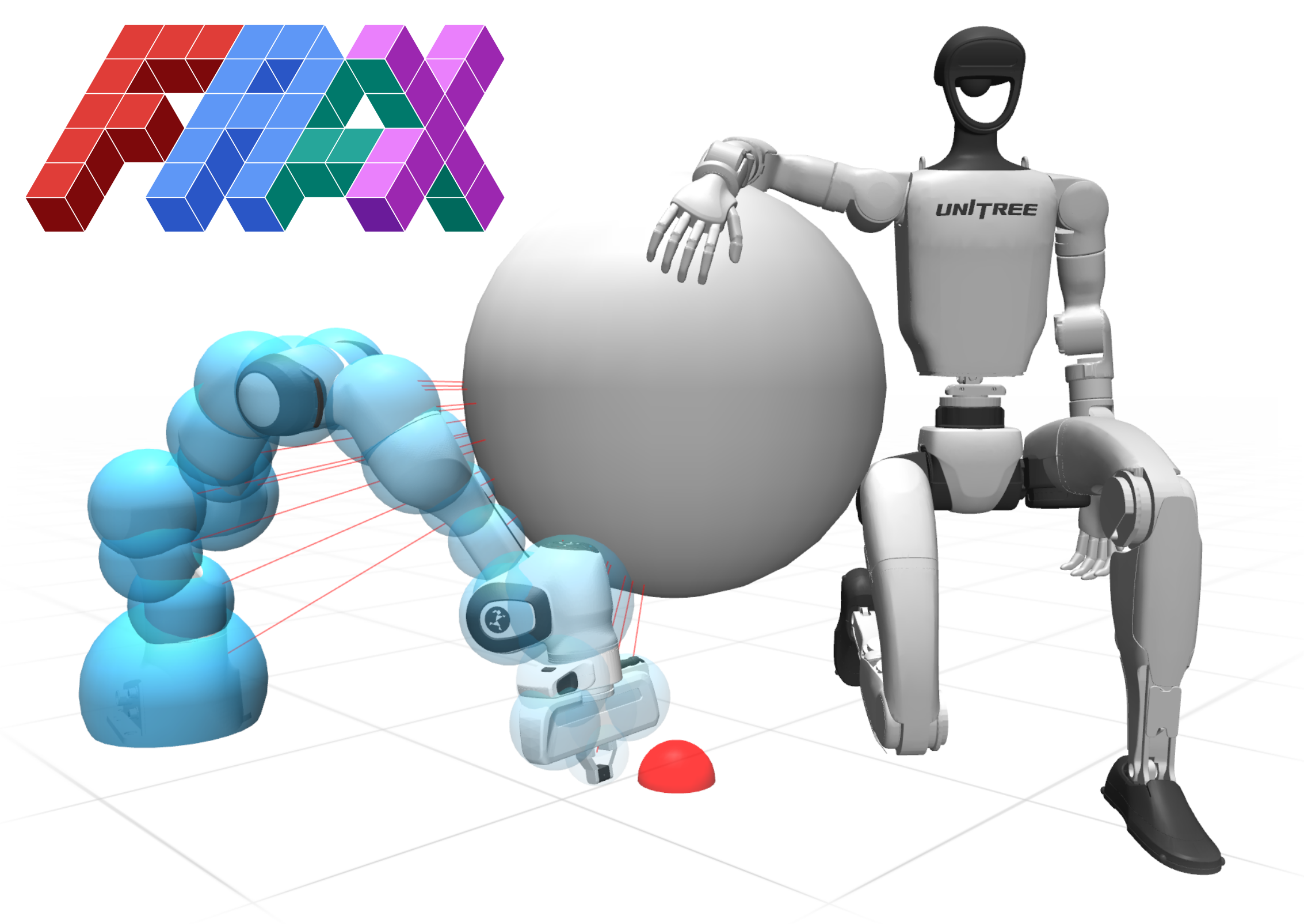}
    \caption{\textbf{\texttt{frax} is a high-performance library for robot kinematics and dynamics on CPU, GPU, and TPU, supporting manipulators, humanoids, and more}. Via JAX, \texttt{frax} enables fast robot control and planning with automatic differentiation through arbitrary functions of the kinematics and dynamics. Shown above (Franka Panda): collision and singularity avoidance in an optimization-based inverse dynamics controller, fully through \texttt{jax.jvp}.}
    \label{fig:frax}
    \vspace{-3mm}
\end{figure}

\begin{table*}[ht]
\smallskip
\centering
\caption{Feature and Compatibility Comparison for Common Dynamics Libraries}
\label{tab:comparison}
\begin{tabular}{@{}lcccccc@{}}
\toprule
                   & CPU   & GPU/TPU & Automatic Differentiation  & Interface & Compilation  &   Best For    \\ \midrule
Pinocchio (C++)    & \cmark  & \xmark   & \xmark\(^\dagger\) & C++       & \cmark           & Fast dynamics and analytical derivatives on CPU\\
MuJoCo (C++)       & \cmark  & \xmark   & \xmark             & C++       & \cmark           & Fast simulation and dynamics on CPU\\
Pinocchio (Python) & \cmark  & \xmark   & \xmark             & Python    & \xmark\(^*\)     & Controller prototyping on CPU\\
MuJoCo (Python)    & \cmark  & \xmark   & \xmark             & Python    & \xmark\(^*\)     & Simulation and controller prototyping on CPU \\
BRAX               & \cmark  & \cmark   & \cmark                & Python    & \cmark           & GPU/TPU-parallelized simulation and RL    \\
MJX                & \cmark  & \cmark   & \cmark                & Python    & \cmark           & GPU/TPU-parallelized simulation and RL   \\
\textbf{\texttt{frax}}               & \cmark  & \cmark   & \cmark                & Python    & \cmark           & Fast controller design on CPU, GPU, and TPU   \\ 
\bottomrule
\addlinespace
\multicolumn{7}{l}{\footnotesize \(\dagger\) Compatible with external AD libraries such as CppAD} \\
\multicolumn{7}{l}{\footnotesize * Python bindings to a compiled library}
\vspace{-3mm}
\end{tabular}
\end{table*}

\section{Kinematics and Dynamics Preliminaries}

We consider a robot modeled as a kinematic tree with generalized coordinates \(\mathbf{q}\). The \textit{forward kinematics} \(FK(\mathbf{q})\) defines the transformation \(\mathbf{T}\) of a link, joint, or other reference frame  \(i\) with respect to the \(0^{th}\) (root) frame, denoted as \({}^{0}\mathbf{T}_i\). \textit{Inverse kinematics} (IK) seeks joint configurations or velocities that achieve a desired task-space pose or motion, either \textit{globally} (solving for \(\mathbf{q}\)) or \textit{differentially} (solving for \(\dot{\mathbf{q}}\)).

\textit{Inverse dynamics} (ID) computes the torque required to achieve a joint acceleration \(\ddot{\mathbf{q}}\) given the joint state \(\left[ \mathbf{q}, \dot{\mathbf{q}}\right]\) and any external contacts or applied forces,
\begin{equation}
    \boldsymbol{\Gamma} = \mathbf{M}(\mathbf{q}) \ddot{\mathbf{q}} + \mathbf{c}(\mathbf{q}, \dot{\mathbf{q}}) + \mathbf{g}(\mathbf{q}) + \mathbf{J}^T(\mathbf{q})\mathbf{f}
    \label{eq:inverse_dynamics}
\end{equation}
Here, \(\boldsymbol{\Gamma}\) is the vector of generalized joint torques, \(\mathbf{M}(\mathbf{q})\) is the joint-space mass matrix, \(\mathbf{c}(\mathbf{q}, \dot{\mathbf{q}})\) is the vector of centrifugal and Coriolis forces, \(\mathbf{g}(\mathbf{q})\) is the gravity vector, and \(\mathbf{J}\) is the Jacobian for any constraints or external forces \(\mathbf{f}\).

The corresponding \textit{forward dynamics} (FD) computes the acceleration of the robot \(\ddot{\mathbf{q}}\) resulting from the applied joint torques and external forces, for a given joint state \(\left[ \mathbf{q}, \dot{\mathbf{q}}\right]\),
\begin{equation}
     \ddot{\mathbf{q}} = \mathbf{M}^{-1}(\mathbf{q})\left[ \boldsymbol{\Gamma} - \mathbf{c}(\mathbf{q}, \dot{\mathbf{q}}) - \mathbf{g}(\mathbf{q}) - \mathbf{J}^T(\mathbf{q})\mathbf{f} \right] 
    \label{eq:forward_dynamics}
\end{equation}

Rigid body dynamics algorithms are commonly based on Featherstone's formulation \cite{featherstone2008rigid}, namely, the Composite Rigid Body Algorithm (CRBA), the Recursive Newton-Euler Algorithm (RNEA), and the Articulated Body Algorithm (ABA). CRBA computes the joint-space mass matrix \(\mathbf{M}\); RNEA computes the inverse dynamics (Eq. \ref{eq:inverse_dynamics}) or individual terms such as \(\mathbf{c}\) or \(\mathbf{g}\); and ABA computes the forward dynamics as a lower-complexity alternative to Eq. \ref{eq:forward_dynamics}.  These methods rely on \textit{spatial algebra}, expressing motion and force in 6-D vector spaces. Using the notation of spatial algebra, we can construct the spatial axes \(\mathbf{S}\), inertias \(\mathbf{I}\), velocities \(\mathbf{V}\), accelerations \(\mathbf{A}\), forces \(\mathbf{F}\), and for CRBA, the composite inertias \(\mathbf{C}\).

\section{Vectorized Rigid-Body Dynamics}

\begin{figure*}[ht]
    \smallskip
    \centering
    \includegraphics[width=\linewidth]{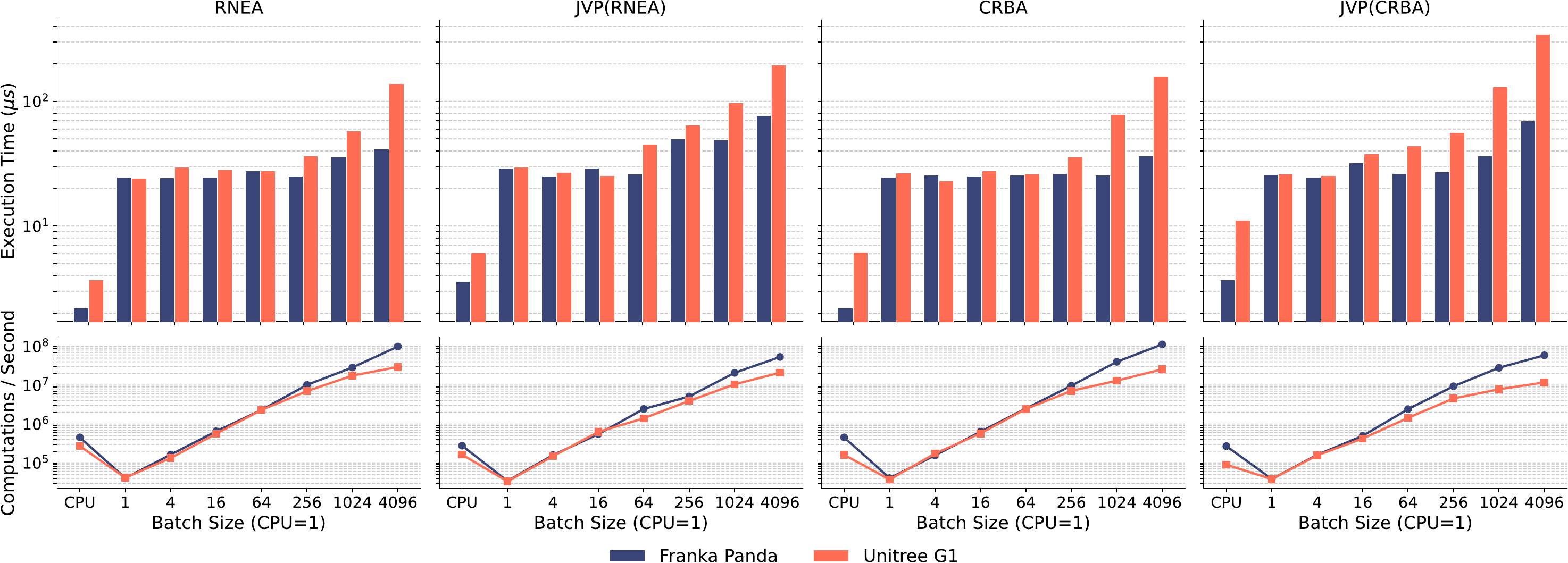}
    \vspace{-3mm}
    \caption{\textbf{Rigid-body dynamics performance across batch sizes}. \texttt{frax}'s vectorized dynamics methods are designed for high performance on CPU and on GPU, including calls to automatic differentiation (\texttt{jax.jvp}) on these methods. All timing values for GPU include any I/O overhead.}
    \label{fig:rbda_scaling}
\end{figure*}

While most dynamics libraries implement these algorithms through a recursive traversal of the kinematic tree, \texttt{frax} instead adopts a fully-vectorized formulation that prioritizes fine-grained parallelism, simplified tracing for automatic differentiation, and fast JIT compilation. Algorithms \ref{alg:vectorized_rnea} and \ref{alg:vectorized_crba} provide an overview of the primary vectorized dynamics methods used in \texttt{frax}. Notation is based on spatial algebra and array broadcasting; a full overview is available in the Appendix (Table \ref{tab:rbda_notation}).

{
\linespread{1.2}\selectfont %
\begin{algorithm}[ht]
\caption{\(\text{Vectorized RNEA}(\mathbf{q}, \dot{\mathbf{q}}, \ddot{\mathbf{q}}, \mathbf{a}_g, {}^{0}\mathbf{F}_{\text{ext}, 1:n}) \rightarrow \boldsymbol{\Gamma}\) }\label{alg:vectorized_rnea}
\begin{algorithmic}[1]
\State \textbf{Precompute: } \(\mathbf{U}\)  \Comment{Ancestor mask}
\State \textbf{Prepare: } \({}^0\mathbf{S}_{1:n}, {}^0\mathbf{I}_{1:n}\) %
\State \(^0\mathbf{V}_{1:n} = \mathbf{U} ({}^{0}\mathbf{S}_{i} \dot{\mathbf{q}}_i)_{i=1}^{n}\)
\State \(^0\mathbf{A}_{1:n} = (\mathbf{a}_g)_{i=1}^{n} + \mathbf{U} ({}^{0}\mathbf{S}_{i} \ddot{\mathbf{q}}_i + {}^{0}\mathbf{V}_i \times {}^{0}\mathbf{S}_i \dot{\mathbf{q}}_i)_{i=1}^{n}\)
\State \(^0\mathbf{F}_{1:n} = \mathbf{U}^T ({}^{0}\mathbf{I}_i {}^{0}\mathbf{A}_i + {}^0\mathbf{V}_i \times^{*} {}^0\mathbf{I}_i  {}^0\mathbf{V}_i - {}^{0}\mathbf{F}_{\text{ext}, i})_{i=1}^{n}\)
\State \(\boldsymbol{\Gamma} = ({}^{0}\mathbf{S}_i \cdot {}^{0} \mathbf{F}_i)_{i=1}^{n}\)
\end{algorithmic}
\end{algorithm}
}

{
\linespread{1.2}\selectfont %
\begin{algorithm}[ht]
\caption{\(\text{Vectorized CRBA}(\mathbf{q}) \rightarrow \mathbf{M}\)}\label{alg:vectorized_crba}
\begin{algorithmic}[1]
\State \textbf{Precompute: } \(\mathbf{U}\) \Comment{Ancestor mask}
\State \textbf{Prepare: } \({}^0\mathbf{S}_{1:n}, {}^0\mathbf{I}_{1:n}\) %
\State \(^0\mathbf{C}_{1:n} = \mathbf{U}^T \ {}^0\mathbf{I}_{1:n}\)
\State \(\mathbf{M}_{\text{lower}} = \mathbf{U} \odot ({}^0\mathbf{S}_{i}^T {}^0\mathbf{C}_{i} {}^0\mathbf{S}_{j})_{i,j=1,1}^{n,n}\) %
\State \(\mathbf{M} = \text{Symmetrize}(\mathbf{M}_{\text{lower}})\)
\end{algorithmic}
\end{algorithm}
}

The main difference between \texttt{frax}'s vectorized approach and traditional recursive approaches comes from the use of an ancestor mask \(\mathbf{U} \in \{0,1\}^{n \times n}\) to encode the tree structure. Specifically, \(\mathbf{U}_{ij} = 1\) if \(j \in \text{Anc}(i)\) (we consider \(i \in \text{Anc}(i)\)), and for a topological sort of the robot joints, \(\mathbf{U}\) is lower triangular. This construction allows for summations up and down the tree to be performed with matrix multiplication, assuming spatial values are expressed in a common reference frame. For CRBA, \(\mathbf{U}\) also allows for vectorized computation of the entries of \(\mathbf{M}\), while masking out any terms corresponding to non-ancestral pairs. Similar operators have appeared in older work, such as \cite{rodriguez1989spatial, fisette1996treelike}, but are uncommon in modern libraries.

\subsection{Complexity Tradeoffs}

Crucially, by avoiding a recursive loop-based structure, we intentionally \textit{increase} the complexity of RNEA from \(O(n)\) to \(O(n^2)\), and from \(O(nd)\) to \(O(n^2)\) for CRBA. Here, \(n\) is the number of degrees of freedom (DOFs) and \(d\) is the maximum depth of the kinematic tree. However, the overhead of redundant \(O(n^2)\) computations is vastly outweighed by the performance gains from the XLA compiler's ability to make use of fine-grained parallelism within the CPU, GPU, or TPU. 

In general, this aversion to loops typically leads to modest improvements to the compute time for a single evaluation on CPU, but significant improvements in JIT compilation, GPU performance, and automatic differentiation performance (on both CPU and GPU). While CPUs can efficiently handle tree traversals, GPUs favor operations with fewer sequential dependencies; thus, we take this approach to maintain strong performance across platforms. And, when applying automatic differentiation to the robot dynamics, the vectorized form avoids the cost of tracing gradients through long unrolled loops, for fast evaluation and compilation.

Note that only one loop is strictly required in \texttt{frax}, to compute the forward kinematics. All dynamics terms are evaluated purely with array broadcasting, as seen in Algorithms \ref{alg:vectorized_rnea} and \ref{alg:vectorized_crba}. And, in the special case of a serial kinematic chain, such as the Franka Panda, the FK loop can even be replaced with an \texttt{associative\_scan}. In this case, both the kinematics and dynamics can be evaluated fully \textit{loop-free}, in the sense that no Python \texttt{for} loops are unrolled in the compiled graph.

\subsection{Batched Performance}

In Fig. \ref{fig:rbda_scaling}, we explore the trends in throughput across CPU and increasing batch sizes on GPU. For a single robot instance, CPU is extremely fast -- much more so than on GPU, which experiences additional overhead cost, though it begins to match the total computations/second achievable once the batch size exceeds 4. As batch size increases, overhead is considerable up to approximately \(N = 64\), after which the increased computational load begins to noticeably impact the total time for the batch. We also see strong performance even on very large batch sizes (4096) with throughput rates of upwards of 100 million computations per second. While this is not directly comparable with a GPU-accelerated simulator's ``steps per second" metric, it does indicate that \texttt{frax}'s dynamics methods can be used in a highly-parallelized context without impeding the performance of an underlying training environment.

\section{Getting Started with FRAX}

After installing either via \texttt{pip} or from source, \texttt{frax} supports any robot loaded via URDF, with specialized methods based on the kinematic structure (end-effector transforms, Jacobians, manipulability indices, ...) available via the Manipulator and Humanoid classes. \texttt{frax} has native support for the Franka Panda and Unitree G1, including tuned spherized collision models and self-collision pairs for both robots.

\texttt{frax} currently contains three examples, with more to be added: differential IK, operational space control, and safe operational space control with CBFs \cite{morton2025oscbf}. Each is a minimal example of designing an IK or ID controller with JAX, and can be fully wrapped in \texttt{jax.jit} for high performance. Figure \ref{fig:frax} gives a visual representation of the CBF demo, including the spherized collision model for the Franka, an obstacle, and the end-effector target shown in red. 

\subsection{Fast JVPs for Robot Control}

As a unique benefit of using JAX, we can often replace a full set of Jacobian computations with an efficient, direct evaluation of a Jacobian-vector product (JVP). In general, a Jacobian computed via AD will be slower than an analytical form\footnote{\texttt{frax} contains analytical Jacobians for many common functions of the robot kinematics, for this reason.}, \textit{unless the Jacobian does not need to be computed in the first place}. As an example, in the CBF demo, we differentiate through the model of the Franka to form the safety constraints, which can be nontrivial functions of the kinematics and dynamics (e.g., the manipulability index). The Lie derivatives required to construct the CBF constraint, such as \(L_fh(\mathbf{z}) = \frac{dh}{d\mathbf{z}}f(\mathbf{z})\), is itself a JVP, and thus \(\frac{dh}{d\mathbf{z}}\) does not need to be computed in full. This can be particularly beneficial for high-DOF systems like the Unitree G1, while also being simple to derive and implement.

\section{A Comparison of Dynamics Libraries}
\label{sec:comparing_libraries}

\begin{figure*}[ht]
    \smallskip
    \centering
    \includegraphics[width=\linewidth]{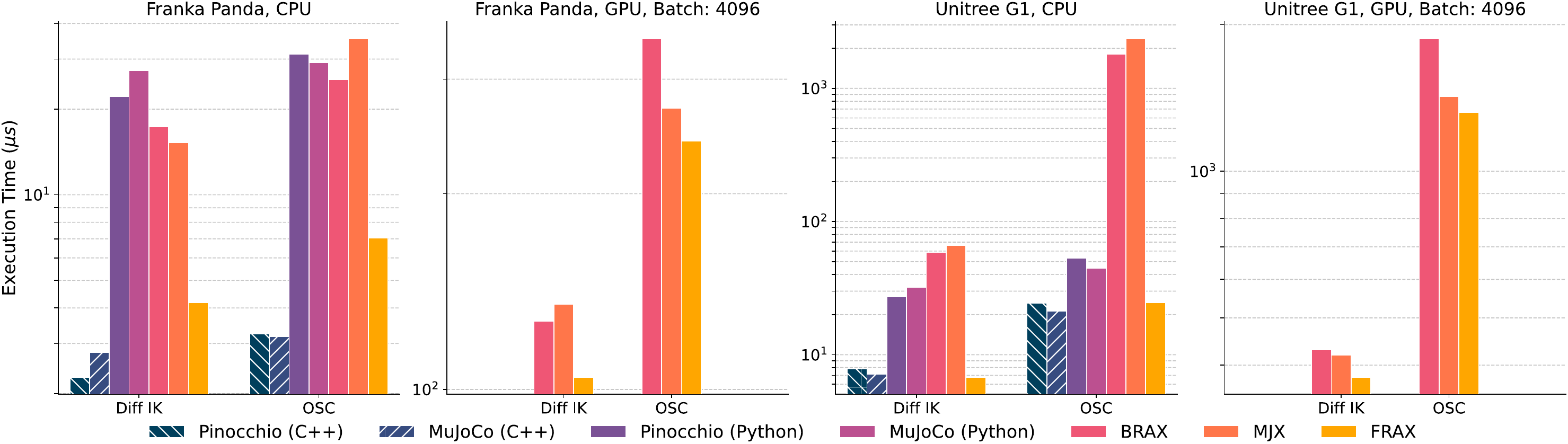}
    \vspace{-3mm}
    \caption{\textbf{Compute timing for controllers written with \texttt{frax} and other common libraries}. On CPU, Python-based controllers are \textit{up to 2-5x faster} with \texttt{frax} than Pinocchio or MuJoCo's bindings, coming close to matching their underlying performance on C++. And, on GPU, \texttt{frax} is on par with MJX or BRAX; enabling high-speed control for \textit{both} single-robot deployment and massively-parallel training.
    }
    \label{fig:controller_timings}
    \vspace{-2mm}
\end{figure*}

To evaluate \texttt{frax}'s performance, we consider the application of writing robot controllers, including IK-based (differential IK) and ID-based (operational space control) methods. For comparison, we include compiled C++ libraries (Pinocchio, MuJoCo), their Python APIs (with NumPy for intermediate computations), and JIT-compiled JAX libraries (MJX, BRAX). These tests reflect a typical workflow, where we need multiple kinematic and dynamic terms (\(\mathbf{M}\), \(\mathbf{c}\), \(\mathbf{g}\), \(\mathbf{J}\), ...).

From Figure \ref{fig:controller_timings}, we observe that existing JAX libraries for robot dynamics (MJX, BRAX) tend to optimize for GPU performance, leading to limited applicability for single-robot control and planning on CPU. This is especially noticeable for inverse dynamics methods on the G1, which can lead to significant spikes in controller compute times (on the order of 1 \(ms\) as opposed to 25 \(\mu s\) with \texttt{frax}). 

In comparison to the Python APIs for Pinocchio and MuJoCo, \texttt{frax} fares very well, with around a 2-5x speedup, due to JAX's JIT compilation leading to reduced overhead, compiler fusions, and multithreaded linear algebra operations.

For the best possible performance on CPU, a controller written in C++ and compiled directly with the C++ interfaces to Pinocchio and MuJoCo will beat out \texttt{frax}, but this gap can be quite close, as seen in the G1 OSC timings (Fig. \ref{fig:controller_timings}). This C++ approach also loses the flexibility of Python, the easy automatic differentiation enabled via JAX, and the ability to run on GPU or TPU (see Table \ref{tab:comparison}). A brief summary of \texttt{frax}'s CPU performance on common IK and ID controllers is also available in Table \ref{tab:cpu_times}, and further discussion on performance is included in the Appendix.

\begin{table}[t]
\centering
\caption{Compute Times for FRAX Controllers on CPU ($\mu$s) }
\label{tab:cpu_times}
\begin{tabular}{@{} lcc @{}}
\toprule
Controller & Franka Panda & Unitree G1 \\
\midrule
Inverse kinematics [Diff. IK] & 4.09 & 6.71 \\
Inverse dynamics [OSC]      & 7.03 & 24.6 \\
\bottomrule
\end{tabular}
\end{table}

On GPU, \texttt{frax} is on par (or slightly faster) than MJX or BRAX. All three were designed for good GPU compatibility with JAX as the backend, so this is to be expected. However, \texttt{frax} has the additional benefit of significantly reduced JIT compilation times --- while JIT times are of much lower importance than ``hot" calls in the control loop,  for these tests, \texttt{frax} compiles in 1-2 seconds versus 6-12 for MJX/BRAX. For iterative development and tuning, this speedup can significantly reduce friction for designers.

\section{Conclusion} 
\label{sec:conclusion}

In this work, we presented \texttt{frax}, a high-performance JAX-based library for robot kinematics and dynamics. Through a vectorized form of rigid-body dynamics, \texttt{frax} is fast enough for real-time, single-robot control on CPU, and scales to thousands of parallel instances on GPU or TPU. Integration with JAX allows users to automatically differentiate through complex functions of the robot kinematics and dynamics, and easily prototype in Python, without sacrificing on performance.

\subsection{Limitations and Future Work}

\texttt{frax} is more minimal than other libraries like Pinocchio and MJX, but this is by design, resulting in high performance on a restricted setting. While C++ can lead to moderately faster control rates (and potentially further improved speed with code generation \cite{carpentier2019pinocchio}), \texttt{frax} provides a much more flexible Python-based AD-compatible interface that is sufficiently fast for kilohertz-rate control on hardware \cite{morton2025oscbf}. And, while MJX can provide a full simulation environment for training RL policies, \texttt{frax} can complement these environments (e.g. a low-level controller for an end-effector action space), and is more flexible for other parallelized applications such as planning. Additionally, \texttt{frax} does not currently focus on solving global IK problems, which is best handled by other libraries \cite{kim2025pyroki}.

Future directions for the library include (as an inexhaustive list) (1) support for collision geometries beyond just spheres and planes, especially capsules, boxes, and ellipsoids; (2) implementations of ABA, analytical derivatives of RNEA, and direct inversion of the mass matrix \cite{carpentier2018analytical, carpentier2019pinocchio}\footnote{Initial experiments indicate that these methods are not well suited to vectorization, and often, it seems to be more practical and faster to just call a Cholesky-based inversion or solve method on the mass matrix.}; and (3) improved support for more classes of robots (e.g. quadrupeds) and more built-in collision models for common robots. And, while we have mainly focused on \texttt{frax}'s application to robot control, the dynamics and automatic differentiation support will be beneficial for a wide range of future planners and nonlinear optimization methods in robotics.

\bibliographystyle{unsrtnat}
\bibliography{references}

\newpage
\appendix
\section{Appendix}

{
\linespread{1.2}\selectfont %
\begin{table}[ht]
    \centering
    \caption{Notation for Algorithms \ref{alg:vectorized_rnea} and \ref{alg:vectorized_crba}}
    \label{tab:rbda_notation}
    \begin{tabular}{@{}lll@{}}
    \toprule
    Notation       & Definition                            & Shape \\ \midrule
    \(\mathbf{U}\)                   & Ancestor mask: \(\mathbf{U}_{ij}=1\) if \(j \in \text{Anc}(i)\) else \(0\)                         & \(n \times n   \) \\
    \(\mathbf{S}\)                   & Spatial axes                          & \(n \times 6   \) \\
    \(\mathbf{I}\)                   & Spatial inertias                      & \(n \times 6 \times 6 \) \\
    \(\mathbf{V}\)                   & Spatial velocities                    & \(n \times 6   \) \\
    \(\mathbf{A}\)                   & Spatial accelerations                 & \(n \times 6   \) \\
    \(\mathbf{F}\)                   & Spatial forces                        & \(n \times 6   \) \\
    \(\mathbf{C}\)                   & Composite inertias                    & \(n \times 6 \times 6 \) \\
    \(\mathbf{a}_g\)                   & Spatial acceleration due to gravity                    & \(6 \) \\
    \(\odot\)               & Hadamard (elementwise) product        & --- \\
    \(\times\)              & Spatial cross product (motion)        & --- \\
    \(\times^*\)              & Spatial cross product (force)        & --- \\
    \((\cdot)_{i=1}^{n}\)                  & Array broadcasting                 & ---   \\
    \({}^{0}(\cdot)\) & Expressed in the root reference frame & ---  \\ 
    \((\cdot)_{\text{lower}}\) & Lower triangular & ---  \\
    \bottomrule
    \end{tabular}
\end{table}
}

\vspace{-2mm}

\subsection{Timing}
\label{subsec:timing}

All timing values were recorded on a desktop PC with an i9-14900KF CPU, 64GB RAM, and an RTX 4090 GPU. Python methods were tested with \texttt{time.perf\_counter} over 10,000 iterations, with \texttt{jax.block\_until\_ready()} called on any JAX functions to avoid timing the asynchronous dispatch. C++ methods were compiled with \texttt{g++ -03} and timed via \texttt{clock\_gettime()} with \texttt{CLOCK\_MONOTONIC} as the source, as per \cite{plancher2022grid}. 

\subsection{JAX Configuration}
\label{subsec:config}

On CPU, JAX is particularly sensitive to versions and configuration flags. We find that the best CPU performance can be achieved by using JAX version 0.4.30 and the following flags: \texttt{JAX\_ENABLE\_x64=1} and \texttt{XLA\_FLAGS=
"--xla\_cpu\_multi\_thread\_eigen=false intra\_op\_parallelism\_threads=1"}. The XLA flags restrict the program to a single thread, which tends to work well for, e.g., robot controllers within a single ROS node. However, this may not be the best choice for all possible workloads with \texttt{frax}, and as such, it is left to the user to specify. 

\subsection{Evaluating Performance}

In general, \texttt{frax} performs best when there are \textit{many} operations (kinematics, dynamics, and otherwise) which can all be captured under a single JIT region, and this is where performance is most directly comparable to C++. In Sec. \ref{sec:comparing_libraries} we use controller design for benchmarking various libraries: this is a representative use-case that requires multiple operations, all of which can be computed with \texttt{frax} and JAX. As such, under JIT, the XLA compiler is able to optimize performance across the full controller evaluation. An alternative benchmark would be to compare performance on a \textit{single} operation such as RNEA or CRBA, but this is a less typical use-case, and overhead would dominate the timing for JAX methods.

\end{document}